\documentclass[12pt,a4paper,twocolumn]{article}
\usepackage{caption} 
\usepackage[utf8]{inputenc}
\usepackage{amsmath} 
\usepackage{amsfonts} 
\usepackage{amssymb} 
\usepackage{graphicx} 
\usepackage[left=2cm,right=2cm,top=2cm,bottom=2cm]{geometry} 
\usepackage{xcolor} 
\usepackage{hyperref} 
\hypersetup{%
  colorlinks=true,
  linkcolor=blue,
  }
 
\usepackage{subcaption}
\usepackage{mwe} 

\author{
  Mohammad M. Haji-Esmaeili\\
  \texttt{mohammadhaji@modares.ac.ir}
  \and
  Gholamali Montazer\\
  \texttt{montazer@modares.ac.ir}
}
\begin{document}

\date{
\includegraphics[width=1.0\linewidth]{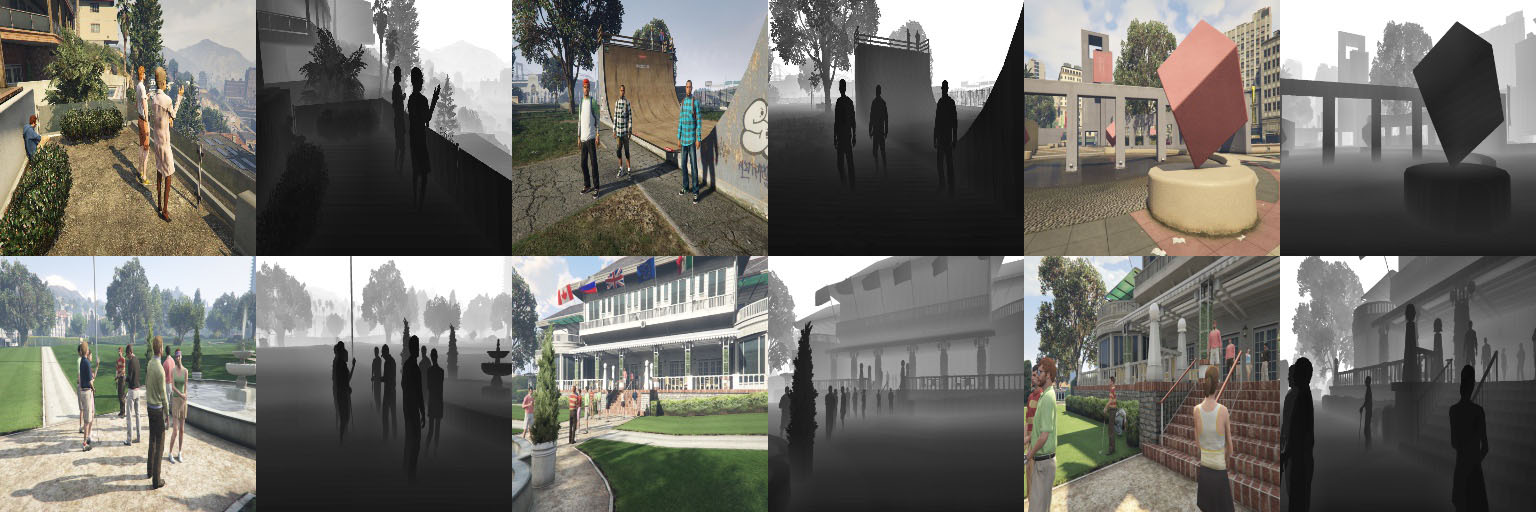}
\captionof{figure}{Images and Depths extracted from the game \mbox{Grand Theft Auto V}}
     \label{fig:dataset-examples}
}
\title{Playing for Depth}
\maketitle

\begin{abstract}

{\it Estimating the relative depth of a scene is a significant step towards understanding the general structure of the depicted scenery, the relations of entities in the scene and their interactions. When faced with the task of estimating depth without the use of Stereo images, we are dependent on the availability of large-scale depth datasets and high-capacity models to capture the intrinsic nature of depth. Unfortunately, creating datasets of depth images is not a trivial task as the requirements for the camera mainly limits us to areas where we can provide  the necessities for the camera to work.

In this work, we present a new depth dataset captured from Video Games in an easy and reproducible way. The nature of open-world video games gives us the ability to capture high-quality depth maps in the wild without the constrictions of Stereo cameras. Experiments on this dataset shows that using such synthetic datasets increases the accuracy of Monocular Depth Estimation in the wild where other approaches usually fail to generalize.}

\end{abstract}

\section{Introduction}

Considering the accurate and fast reaction of human visual cortex in estimating the depth of a scene, the task of Monocular Depth Estimation is still an open problem which is more complex than trying to estimate the depth using pairs of stereo images. contrary to their more straightforward approach, Stereo correspondence requires the availability of more than one image to estimate the depth of a scene which is not feasible in many scenarios \cite{Hartley2003} \cite{Scharstein2001}. Limited availability of Stereo cameras and their rare use in everyday activities are what drives the researchers towards estimating depth using only monocular images.

Estimating monocular depth is a challenging area of research as it is tied to both understanding the global structure of an image and the relations of its entities with each other. For humans who have a rich understanding of the world around them this makes for a trivial task to estimate depth using Monocular Cues such as Textures, Occlusions \& Haze, Brightness, Defocus \& Blur, Interpolation and Linear Perspective \cite{Saxena2006}.

\begin{figure*}
    \centering
    \includegraphics[width=1.0\linewidth]{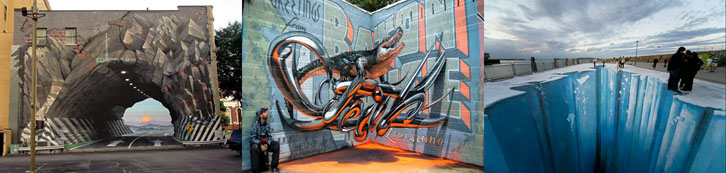}
    \caption{Optical Illusions based on fooling the Monocular Cues}
    \label{fig:optical-illusions}
\end{figure*}

\begin{figure*}
    \centering
    \includegraphics[width=1.0\linewidth]{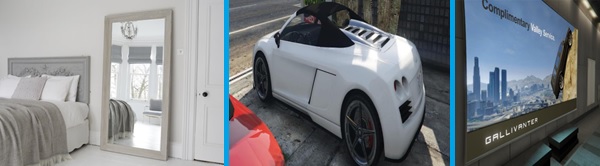}
    \caption{Misleading situations for Depth Estimation in the wild}
    \label{fig:misleading-situations}
\end{figure*}

Integrating the aforementioned cues to end up with a monolithic model capable of estimating depth is a challenging task. Considering all of the monocular cues in a scene, it is still possible to conjure a situation in which a human is mislead about the depth of the scene. There are a variety of situations in the wild in which the global structure of the scene is misleading enough to lead the model to estimate the depth completely wrong. These situations can be created artificially using Optical Illusions (Fig \ref{fig:optical-illusions}) or naturally occur in the wild (Fig \ref{fig:misleading-situations}) where the only cue helping us estimate the depth is the context of the scene as the Monocular Cues are rarely of help considering the complexity of these scenes.

Some instances where the scene is illusive enough to result in estimation errors are shown in Figure \ref{fig:misleading-situations}: Lacking the necessary knowledge of how Mirrors look and work results in trying to estimate the depth of the mirror's reflection. Transparent objects such as a car's windshield or an apartment's glassy windows results in errors when estimating depth. an advertisement or poster printed on a wall should only be treated as a flat object with linear and flat depth which is hard for most models to understand. These are examples of situations where the traditional methods of depth estimation \cite{Shao1988} fail to handle the complexity of scenes.

The recent achievements in Depth Estimation is widely attributed to the advancements of deeper architectures and large scale datasets used to train them. Unfortunately the scope of locations these datasets provide are lacking in variety, for instance notable depth datasets are limited to where the capturing camera could roam free such as interiors \cite{Silberman2012} or roads \cite{Geiger2013}. Extrapolating for scenes which are outside of these domains occasionally results in inaccurate depth estimations. On the other hand a large portion of these datasets contain heavy noises which forces the researchers to either mask them completely or fill them using inpainting \cite{Eigen2014} \cite{Eigen2016}.

The limited scope and noise-ridden state of these datasets has resulted in attempts to gather depth datasets using other means than Stereo images \cite{Xie2016} \cite{Chen2016a}. In this paper, we use a novel approach to create a depth dataset which can contain a vast array of different locations in the wild with no noise whatsoever. We then introduce an architecture minimizing multiple losses to estimate the depth of images in the wild and provide comparisons with other notable works in the field.

\section{Related Work}

\noindent
\textbf{User-Specified Depth Inputs.} By considering a scene as a combination of Foreground and Background and having the aid of users to determine these two layers, one can estimate the global depth of a scene \cite{Iizuka2014}. The main problem of these approaches is the need of user hints about the relative depth of the scene without which the estimation would not be able to work.

\noindent
\textbf{Monocular Datasets.} Until recently a great deal of research in depth estimation was based on using a handful of RGB-D datasets which were mainly limited to scenes where the capturing device\footnote{Mainly Kinect or LIDAR devices} could freely roam, e.g. indoor scenes \cite{Silberman2012} or constrained outdoor locations \cite{Saxena2006}. The main problems of these datasets are the location limitations, noisy captures which need inpainting or masking and the time consuming procedure of physically moving the camera around by hand or by using a robot.

\noindent
\textbf{Ordinal Datasets.} Instead of trying to label full-scene depth images, these approaches rely on point-wise ordinal relations to estimate the depth of the scene \cite{Zoran2016} \cite{Chen2016a} \cite{Ma2017}. The dataset is gathered mainly by crowd-sourcing and the goal is for each user to differentiate between the relational depth of two random points in the scene. By having lots of points that can tell which objects are farther and which ones are closer, a model can be built to approximate the global ordinal depth of the scene. a hardship specific to this approach is the method of data gathering which relies mainly on crowd-sourced monetary services such as Amazon Mechanical Turk.

\noindent
\textbf{View Generation.} This approach uses a collection of images from different viewpoints of a scene to generate a new view for the scene \cite{Flynn2015}. This idea has been used on Side-by-Side 3D Movies (as a dataset) to generate the right image from the left image \cite{Xie2016} \cite{Bae2017}. This process generates probabilistic disparity maps which can then be used to shift the left image (as input) and render the right image (as output). One advantage of 3D Movies is the vast number of frames which can be extracted from the videos and used as a dataset. Another way to approach this method is to use many Internet photos to approximate a 3D model of a scene \cite{Li2018}. By gathering a large dataset of real-world landmarks with many photos for each one, one can construct a 3D Model of the landmark using SfM\footnote{Structure from Motion} and MVS\footnote{Multi View Stereo} methods. After refining the depth maps given by the reconstructed 3D scene, one can use methods from previously discussed work (such as Monocular \& Ordinal methods) to estimate a final depth for the image. The depth maps aid in general depth estimation while the point clouds are used for determining Ordinal relations between objects and both contribute to the final Depth of the scene \cite{Li2018}. Reliance on SfM and MVS methods takes many steps to finally end up with a proper dataset to train on. a State-of-the-Art application should estimate the 3D structure which are usually noisy in nature, a different step should be taken to refine the noisy data and correct/remove outliers (such as people standing in front of a landmark). Our approach does not need these refinement and preprocessing steps, as the depth maps are properly rendered and no noise is present in their structure.

\section{Dataset Construction}

\begin{table*}[ht]
\centering
\begin{tabular}{c|c|c|c}

\rotatebox{90}{Limited Datasets}&
\includegraphics[width=0.3\textwidth]{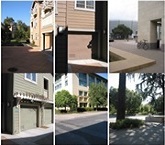}&
\includegraphics[width=0.3\textwidth]{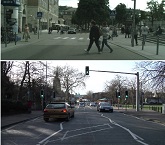}&\includegraphics[width=0.3\textwidth]{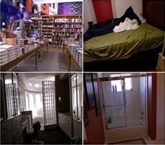}
\\

&
Make3D&
Kitti&
NYU
\\

\hline
\rotatebox{90}{Unlimited Datasets}&
\multicolumn{3}{c}{
\includegraphics[width=0.45\textwidth]{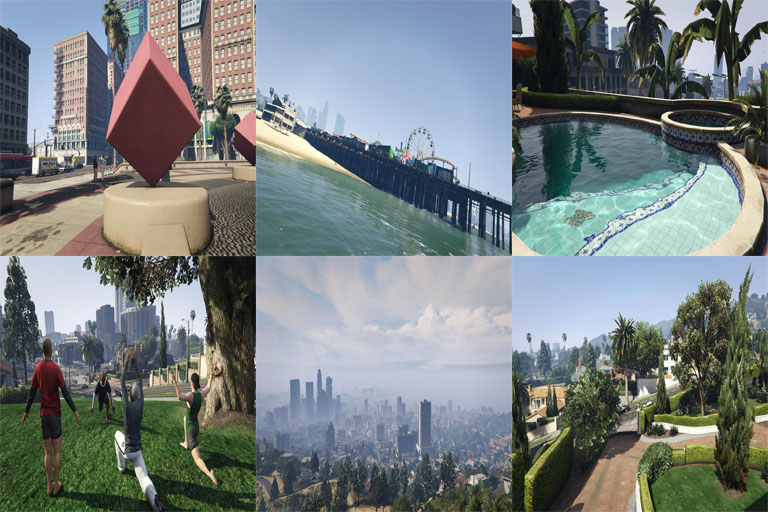}
\includegraphics[width=0.45\textwidth]{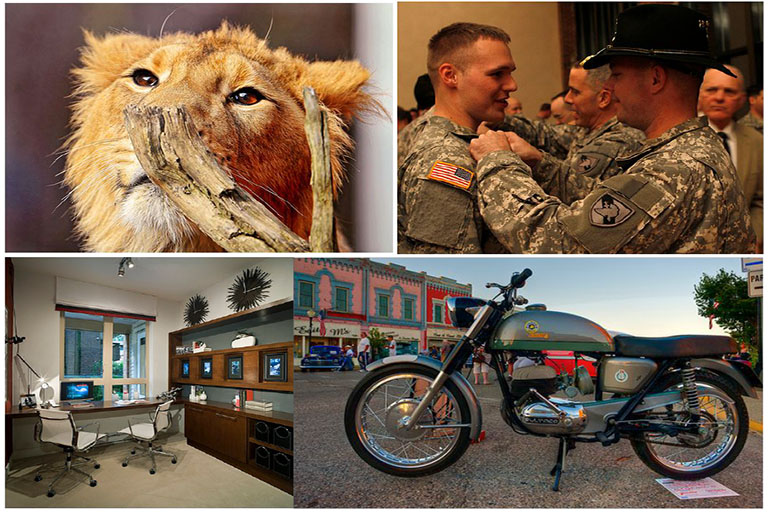}
}
\\

&
\multicolumn{3}{c}{
\makebox[0.5\textwidth][c]{Our Dataset}
\makebox[0.5\textwidth][c]{Depth in the Wild}
}

\end{tabular}
\captionof{figure}{Notable Depth datasets}
\label{img:notable-depth-datasets}
\end{table*}

For the past couple of years Video Game realism has had a steady growth thanks to a vast array of complex technological advancements and raw computational power.  On the other hand the world scale of these game have also been rising both in quantity and quality\footnote{e.g. Just Cause 3 has a map size of 627 $km^2$ and The Witcher 3 has a map size of 33 $km^2$}. Even considering the large scale of these maps, the details which fill these worlds and give them their lifelike characteristics can severely differ from each other\footnote{e.g. while Final Fantasy XV has a map size of 1126 $km^2$, most of it is not traversable and can only be flown over}. Considering this we have chosen Grand Theft Auto V as the main source of generating samples for the Gaming for Depth (GfD) dataset. The vast variety of places, events, actions and objects in this game is astounding and the attention to detail in this game is superior to most games. Interior details, streets full of people and vehicles, residential and recreational areas, different building architectures, different and controllable weather and lighting conditions are just a small fraction of the lively world of GTA V. These promising characteristics make this game a suitable candidate for automatic extraction of hard-to-acquire real world datasets for tasks such as Semantic Segmentation and Depth Estimation.

To extract the depth maps, We inject a DirectX driver to the game to redirect all of the rendering commands to it. We then proceed to extract the Depth info of each frame displayed within the game and record it. 
This gives us the ability to capture depth frames up to 8K resolutions.

The nature of video games gives us the ability to meddle with the world around us. This is a powerful ability for dataset generation since one can come up with different lighting or situations for just one simple scene. Most of the Depth datasets don't have depth data for situations where the overall lighting may be too dim or even don't have enough lighting variety for one particular scene. Most models based on these datasets end up functioning normal if presented with the same lighting conditions and they may act unfavorably when met with input which has a different light setting. To avoid this problem we inject a trainer into the game to control the global lighting conditions by timing the speed of Sunrise and Sunset. Just by accelerating this timeframe we can have a series of depth images from one scene with different light settings, e.g. sunrise, midday, sunset and midnight.

We also account for different weather conditions by injecting another trainer to control the weather and automatically changing it in short intervals. This gives us the ability to extract different samples from weather conditions such a sunny, rainy, foggy, stormy and smoggy. Some samples of the dataset showing day/night cycles and weather conditions are shown in section \ref{section:appendix}.

The GfD dataset consists of images scaling from half a meter to several Kilometers long. This property of the dataset makes it hard to test the accuracy of the models trained on it with limited-scope indoor datasets such as NYU or SunRGBD. Training a model naively on the GfD dataset results in nearly the same overall depth for the mentioned datasets since most of the scenes are limited to indoor distances. One way to reduce the effects of this range, is to preprocess and normalize the images before training. We have tested Histogram Equalization, Log-Transforms and Standardization techniques to normalize the range of the depth maps. Histogram Equalization, results in depth maps in which the depth of the scene is localized between the nearest and farthest objects, meaning that we end up with a "relativistic" depth between the objects and not the depth that was originally extracted from the game (a characteristic needed in some use-cases and disregarded in others). a downside to Histogram Equalization is the amount of noise it introduces to the overall image especially in scales smaller than the initial image resolution. The log-transform of the depth maps proved to be more useful in cases which we want the overall depth of the scene to stay relevant after rescaling it. 
Standardizing the images proved to be effective where the former techniques failed. There are various filters applied to the user's viewpoint throughout the game. These filters are used when wearing shaded glasses or helmets, in different weather conditions or in some cutscenes of the game to give the overall image a specific color tone. Standardizing the images removes the effect of these color tones and makes the  image colorfully relatable to the rest of the images in the dataset.

We used the Histogram-Equalized, Log-Transformed and Standardized versions of the GfD dataset as inputs to the model and for the sake of global depth scale, the Standardized version worked better considering the propagation problem of brightness intensity (referred in \ref{subsection:network-architecture}) while the Histogram-Equalized version proved useful in vision tasks where we only needed a relative notion of depth between the objects in the scene.

\section{Method}

This section explains the learning architecture and the Loss objectives used to model the Depth Estimation task.

\subsection{Network Architectures}  \label{subsection:network-architecture}

Our initial experiments with the GfD dataset consisted of networks that had worked properly with depth data in the recent literature, which frequently involves the variations of the U-Net (Hourglass) \cite{Chen2016a} \cite{Li2018} and Inception \cite{Szegedy2015} architectures. One of the driving forces behind this decision is the ability of these architectures to prevent the loss of high frequency details in the subsequent layers of the network thanks to the skip-like layers. In the U-Net architecture, the middle layer also acts as an embedding layer for the domain of training images. The combination of the middle embedding layer with the skip layers are effective in Image-to-Image tasks where we need to transform an image into another domain, e.g. Automatic Colorization\cite{Isola2016}, Scene Rendering\cite{Wang2017b}.

These specific characteristics of the U-Net are what makes it rather unsuitable for the use case of image-to-image Depth Estimation. Unlike tasks such as Image Colorization where the skip layers positively affect the generation of global and local details in the predicted image, these layers tend to negatively impact the prediction in the task of Depth Estimation.

The outer skip layers of the U-Net are able to bypass the convolutions taking place in the inner layers and be present at the final layers of the network. While this is a powerful characteristic in many tasks, it usually backfires while estimating depth. The presence of initial layers of the network at the final layers, poses the risk of brightness intensity saturation in which the network prefers to estimate the depth solely based on the intensity data provided by the initial skip layers of the U-Net.

\begin{figure}[t]
    \centering
    \includegraphics[width=1.0\linewidth]{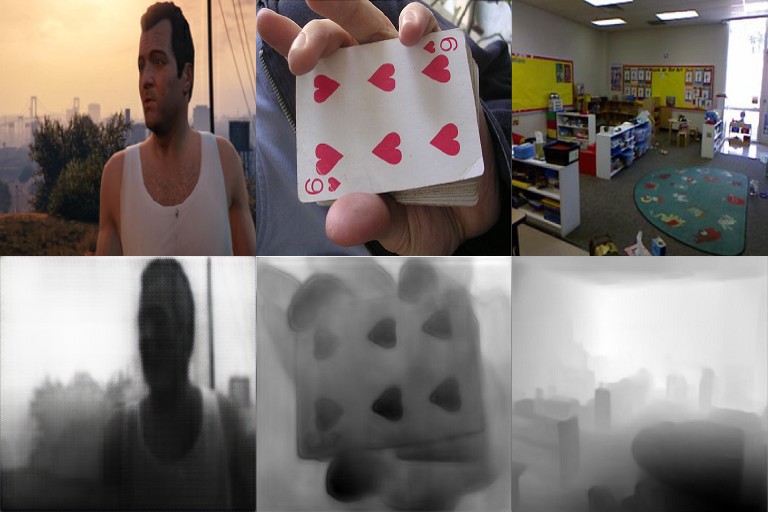}
    \caption{The effect of overusing Brightness Intensity for Depth Estimation.}
    \label{fig:brightness-problem}
\end{figure}

\begin{figure}[t]
    \centering
    \includegraphics[width=1.0\linewidth]{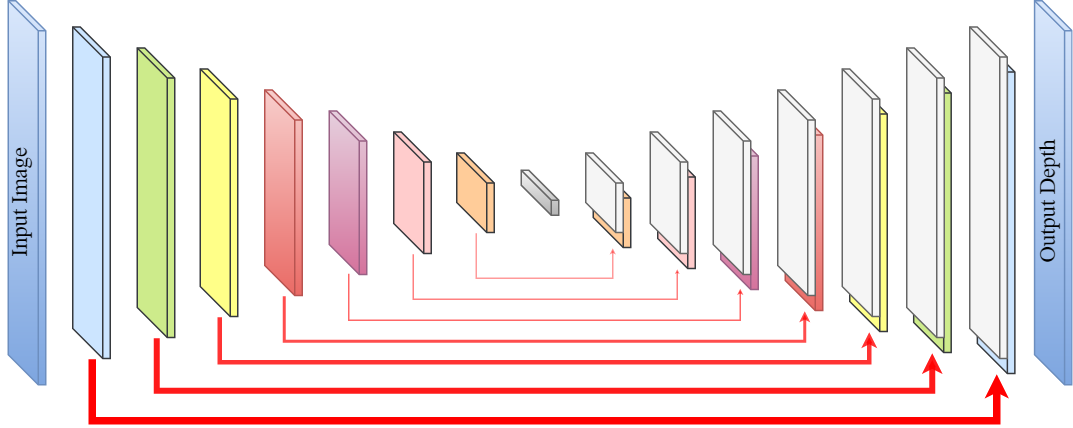}
    \caption{The effect of outer skip layers on the network's reliance on brightness intensity.}
    \label{fig:unet-with-dropout}
\end{figure}

This problem is presented in Figures \ref{fig:brightness-problem} and \ref{fig:unet-with-dropout}. Without the aid of skip layers, generating Depth maps and refining them would be hard, but while used, they result in cases where the network relies heavily on the data provided by them. The shirt worn by the man in Fig \ref{fig:brightness-problem} should be considered to have the same depth as him, which also applies to the hearts on the card \cite{Chen2016a} and the carpet on the ground \cite{Chen2016a}. This problem is present in many situations such as colorful paintings on a white wall, symbols and text on objects or an open window inside a dark room: In all these situations the network prefers to estimate the depth based mainly on the brightness intensity and not the intrinsics of the image, resulting in subpar performance in many situations. Applying weights and dynamic Dropouts to outer skip layers does not hinder their negative effects on the final depth map.

Another family of architectures which showed promising results in our experiments were the networks designed for the task of Semantic Segmentation \cite{Chen2017} \cite{Chen2018}. Designed for classifying individual pixels, these models exhibit stronger inference between objects of the scene when faced with the task of Depth Estimation. One downside to these architectures are the rather large memory requirements which forced us to use at most 3 batch sizes at once on a Geforce 1080Ti. Another problem is the noisy artifacts introduced during training to the final depth maps. Even using the multi-scale scale invariant gradient matching objective (explained in section \ref{subsection:objective-functions}) couldn't completely rid the outputs of these networks from these noisy estimation.

These limitations led us to design a residual architecture which is able to account for the downsides mentioned above.

\subsection{Residual Architecture}

With the recent advent of Residual networks and their outstanding performance in Computer Vision tasks, we have seen a number of tries to apply these architectures to the task of Depth Estimation \cite{Li2017b} \cite{Laina2016}. Depth Estimation is inherently a harder task than other image to image translation topics such as colorization, style transfer or scene rendering due to the need for inference between the object relations in the scene. To maximize the capacity of our model, we opt to use a modified version of Resnet in which Batch Normalization and subsequent ReLU activations have been stripped from each residual block \cite{He2016}.  

Based on a series of experiments, we empirically found removing the Batch Normalization layers beneficial to the overall quality of estimated depth maps and prevent the creation of noisy artifacts after prolonged training of the model. This issue along with the memory consumption of the batch normalization layer (which uses the same amount of memory as the previous convolutional layer), led us to remove this layer as also done in \cite{Nah2017}. a Comparison of our residual block and the original implementation \cite{He2015a} is shown in Fig \ref{img:origial_residual_blocks_and_ours}.

\begin{table*}[ht]
\setlength{\tabcolsep}{32pt}
\centering
\begin{tabular}{cc}

\includegraphics[width=0.25\textwidth]{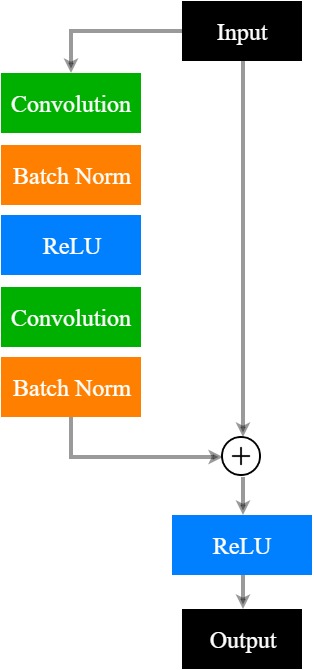}&
\includegraphics[width=0.35\textwidth]{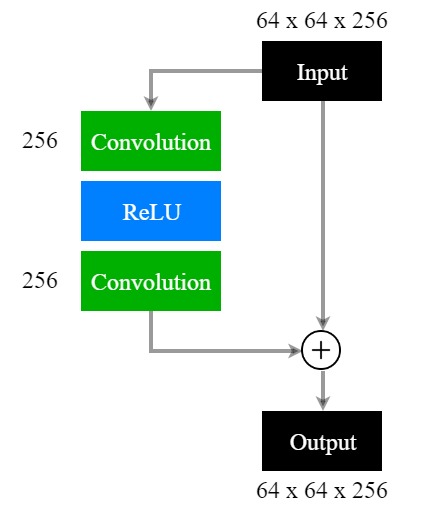}

\\
(a) Original&
(b) Ours

\end{tabular}
\captionof{figure}{Difference between the original Residual Blocks \cite{He2015a} and ours}
\label{img:origial_residual_blocks_and_ours}
\end{table*}

We input a $256\times256$ image to the network and downsample it to $64\times64$ dimensions. We then use 32 residual blocks of 64x64 input sizes where each block has 256 convolutional layers. We use the subpixel convolutional neural network layer proposed in \cite{Shi2016a} to upsample the final residual block's output to $256\times256$. This upsampling technique introduced lesser artifacts and pixelated outputs to the final depth map where the bilinear and transposed convolution layers fail or produce noisy artifacts.

\subsection{Objective Functions}
\label{subsection:objective-functions}
\noindent
\textbf{Scale-Invariant Loss.} The depth range of the GfD dataset makes comparing the predicted depth maps with the ground truth impractical. To solve this problem we use the approaches introduced in \cite{Eigen2014} \cite{Eigen2016}. Even if the scaling of depth maps are different, the relative depth of pairs of corresponding pixels between the predicted depth map $y$ and ground truth $y^{*}$ should stay the same. This relation can be defined in the log-domain as:

{\footnotesize
\begin{equation}
\mathcal{L}_{SI} = \frac{1}{2n^2}\sum\limits_{i,j}((\log y_{i} - \log y_{j}) - (\log y^{*}_{i} - \log y^{*}_{j}))^2
\label{eq:scale-invariance-1}
\end{equation}
}

Expanding and simplifying equation \ref{eq:scale-invariance-1}  results in an equation with similarities to the $l2$ (and also \textit{MSE}) loss functions (let $D_{i}=\log y - \log y^{*}$):

{
\begin{equation}
\mathcal{L}_{SI} = \frac{1}{n}\sum D_{i}^2 - \frac{1}{n^2}(\sum D_{i})^2
\label{eq:scale-invariance-2}
\end{equation}
}

\noindent
\textbf{Total Variation Loss.} One of the notable aspects of depth maps is their usual continuity in local patches of the image. When training a network on the task of pixel-based image to image translation, we are faced with discontinuities and noisy artifacts in the overall predicted depth map.

Relying mainly on the L2 or Scale-Invariant loss functions proves effective for estimating the global depth of the scene but it usually fails to account for the small high-frequency details in the scene. Minimizing the L2-based loss functions is the same as maximizing the log-likelihood of a Gaussian, thus assuming our Depth images come from a Gaussian distribution. This assumption which is a fundamental part of our loss functions, results in the blurry depth predictions of our network.

There are different approaches to improving the quality of the predicted blurry depth maps. Using Conditional GANs\footnote{Generative Adversarial Networks} \cite{Isola2016} is a relatively recent approach in improving the quality of the depth images. The downside of this method is the possible corruption of the generated depth images. The Discriminator is trained to discriminate between the real and fake images generated by the Generator. Because of a lack of context, the Discriminator only tries to make the generated images "look" more realistic without any access to the context of the task which results in depth images which look real but are fundamentally flawed in many semantic concepts related to Depth Estimation.

We use the Total Variation loss as an objective function to enforce the matching of ground truth gradients with that of the predicted depth map. This term not only enforces the gradients and edges to be similar to the ground truth but also contributes to smoothing the overall prediction and controlling the amount of depth discontinuities and noisy artifacts in the final output:

{
\begin{equation}
\mathcal{L}_{TV} = \frac{1}{n}\sum[(\triangledown_{x}D_{i})^2 + (\triangledown_{y}D_{i})^2]
\label{eq:total_variation_single_scale}
\end{equation}
}

While effective for a single scale, we found that applying this term to multiple scales of the generated output can improve the quality of the predicted depth maps even further:

{
\begin{equation}
\mathcal{L}_{TV} = \frac{1}{n}\sum[(\triangledown_{x}D_{i}^{s})^2 + (\triangledown_{y}D_{i}^{s})^2]
\label{eq:total_variation_multi_scale}
\end{equation}
}

where $D_{i}^{s}$ is the depth map value at scale $s$ and position $i$.

This criterion has been used successfully in tasks where we want the generated image to stay consistent in areas with the same intensity such as Colorization \cite{Levin2004} and Depth Estimation \cite{Eigen2016}. Fig \ref{fig:effect_of_tv_loss} demonstrates the effect of properly using Multi-scale Total Variation loss as an objective function:

\begin{figure}[!ht]
    \centering
    \includegraphics[width=1.0\linewidth]{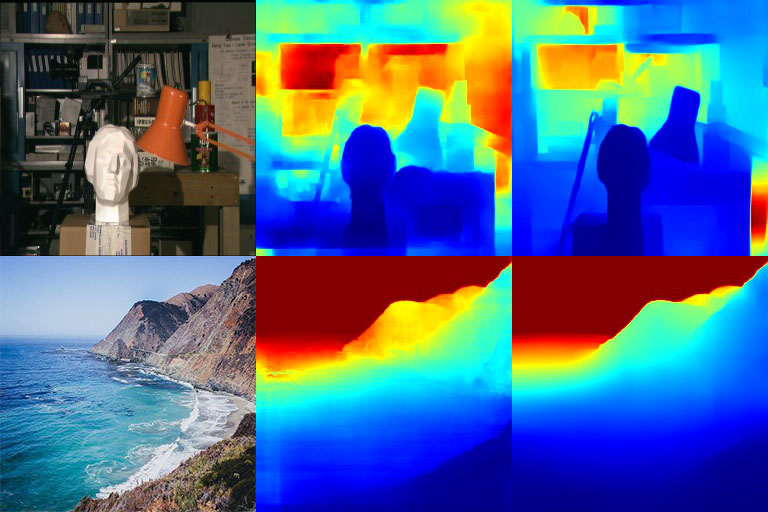}
    \caption{Effect of Multi-scale Total Variation loss on the model. From left to right: Input Image, Without TV loss, With TV Loss}
    \label{fig:effect_of_tv_loss}
\end{figure}

The final objective function is a combination of the aforementioned functions:

\begin{equation}
\mathcal{L}_{Total} = \mathcal{L}_{SI} + \alpha\mathcal{L}_{TV}
\end{equation}

\section{Training \& Evaluation}

We train the model for 50 epochs and use the ADAM optimizer with a learning rate of $0.0004$ and its default parameters. We halve the learning rate by $0.65 \times LR$ every $10^5$ iterations as this allows the network to settle on a final local optimum. We train using batches of size 16. We only train the model on the GfD dataset and do not train the model on any of the test datasets (which results in differences of depth scale).

We apply random scaling to images (up to 30 pixels) and crop the scaled version back to 256 pixels. We also flip the batches randomly to potentially double the dataset capacity.

We tested the final model on the Make3D and NYUDepth v2 datasets. We compare our method to prior works of \cite{Eigen2016} and \cite{Chen2016a}.

\begin{table}[!h]
\centering
\captionof{table}{NYU v2 Benchmarks}
\label{table:benchmarks-NYU}
\begin{tabular}{|c|c|c|c|c|c}
\hline 
Method & RMSE & \vtop{\hbox{\strut RMSE}\hbox{\strut \,\,\,(log)}} & \vtop{\hbox{\strut RMSE}\hbox{\strut \,\,\,(SI)}} & absrel & sqrrel \\ 
\hline 
Eigen (A) & 0.75 & 0.26 & 0.20 & 0.21 & 0.19 \\ 
\hline 
Eigen (V) & 0.64 & \textbf{0.21} & \textbf{0.17} & \textbf{0.16} & \textbf{0.12} \\ 
\hline 
Wang & 0.75 & - & - & 0.22 & - \\ 
\hline 
Liu & 0.82 & - & - & 0.23 & - \\ 
\hline 
Li & 0.82 & - & - & 0.23 & - \\ 
\hline 
Karsch & 1.20 & - & - & 0.35 & - \\ 
\hline 
Baig & 1.0 & - & - & 0.3 & -\\ 
\hline
Zoran & 1.20 & 0.42 & - & 0.40 & 0.54 \\ 
\hline 
Chen & 1.13 & 0.39 & 0.26 & 0.36 & 0.46 \\ 
\hline 
Chen (Full) & 1.10 & 0.38 & 0.24 & 0.34 & 0.42 \\ 
\hline 
\textbf{Ours} & \textbf{0.51} & 0.28 & 0.25 & 0.22 & 0.31 \\ 
\hline 
\end{tabular} 
\end{table}

\subsection{Qualitative Results}

\begin{figure*}[!ht]
    \centering
    \includegraphics[width=1.0\linewidth]{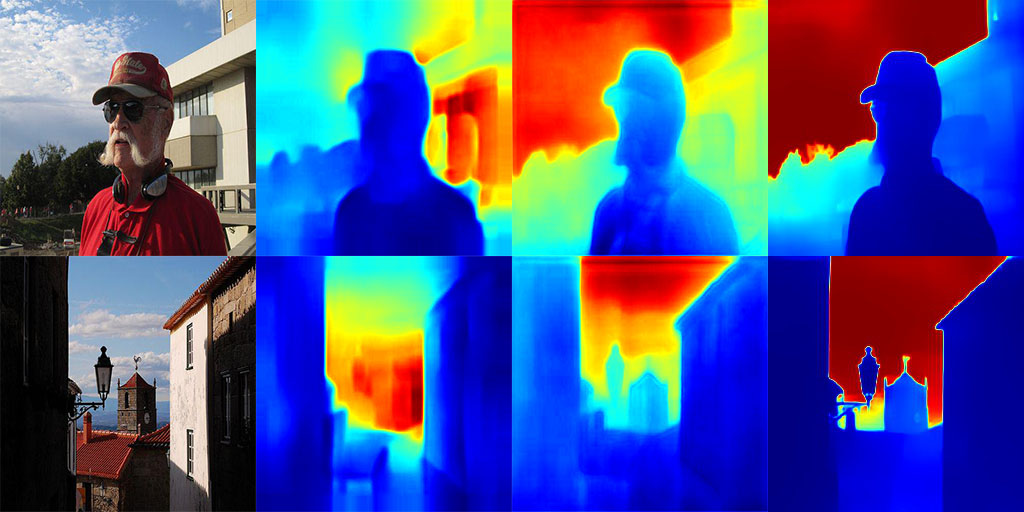}
    \caption{Qualitative Results. From left to right: Input Image, Eigen \cite{Eigen2016}, DIW (Full) \cite{Chen2016a}, Our Approach}
    \label{fig:qualitative_results_brief}
\end{figure*}

Fig \ref{fig:qualitative_results_brief} contains some qualitative comparisons of our model with other notable aproaches on the DIW \cite{Chen2016a} dataset (a more detailed comparison is given in the appendix).

\begin{figure}[!ht]
    \centering
    \includegraphics[width=1.0\linewidth]{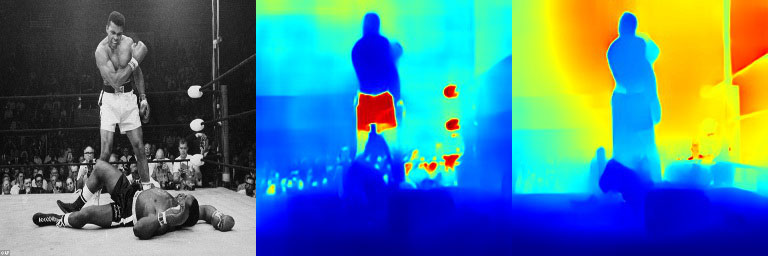}
    \caption{Effect of Standardization on the model. From left to right: Input Image, Normal GfD Dataset, Standardized GfD Dataset}
    \label{fig:effect_of_standardization}
\end{figure}

Standardizing the dataset has resulted in resistance to the brightness intensity and "white" colored objects. This has been demonstrated in Fig \ref{fig:effect_of_standardization}.

\subsection{Failure Cases}
\begin{figure}[!ht]
    \centering
    \includegraphics[width=1.0\linewidth]{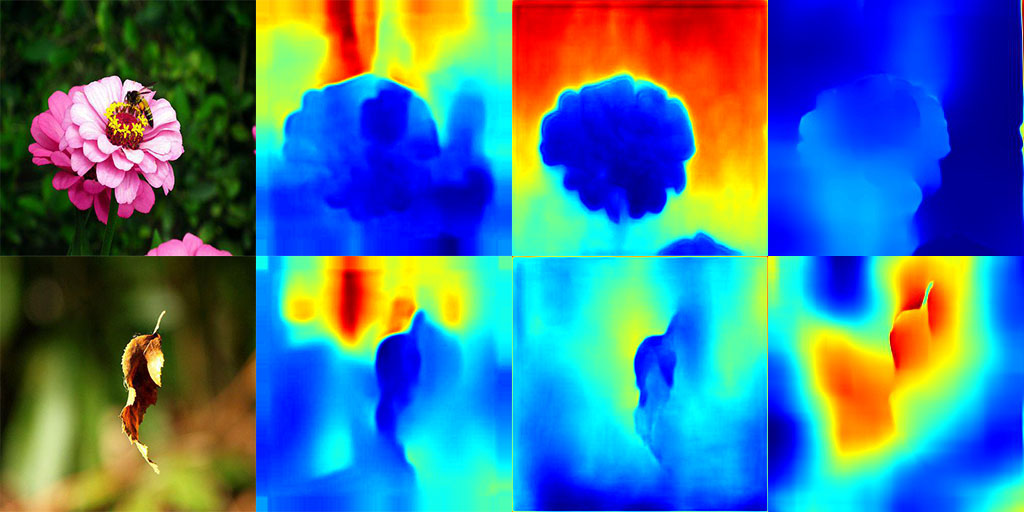}
    \caption{Failure Cases. From left to right: Input Image, Eigen \cite{Eigen2016}, DIW (Full) \cite{Chen2016a}, Our Approach}
    \label{fig:failure_cases}
\end{figure}

We disabled the Depth of Field and Motion Blur settings in GTA V to be able to capture high definition images from the game. This choice results in a dataset which has little to offer to the model in case of depth information related to blurred and out of focus images. The consequence of this choice can be seen in images where the majority of the scene is out of focus and blurred. Fig \ref{fig:failure_cases} contains some of these cases where the model fails to comprehend the global structure of the scene.

\section{Conclusions}

In this paper we have shown a new way to gather depth images from Video Games and use them to train models which are useful in the wild and not limited to the current domain-limited datasets. This approach is easy to use and can easily be done with a vast array of different Video Games available on the market. We then used this dataset of nearly $200,000$ images to train a model which achieves comparable performance in the task of Depth Estimation in the Wild.

The idea of using the extracted Depth of Video Games can be extended to use the other semantic information which lies dormant in the Video Game data such as the Semantic Info of the scene (used for semantic segmentation) \cite{Richter2016} \cite{Richter2017} and also the Normal Surface Maps of the scene. These two auxiliary information can greatly help in estimating the Depth of a scene as shown in \cite{Eigen2016}. In future we hope to expand the GfD dataset to more auxiliary information which can then be used to construct a model which has access to these data for better prediction in the real world scenarios.

\clearpage
\onecolumn
\section{Appendix} \label{section:appendix}

\begin{figure}[!ht]
    \centering
    \includegraphics[width=1.0\linewidth]{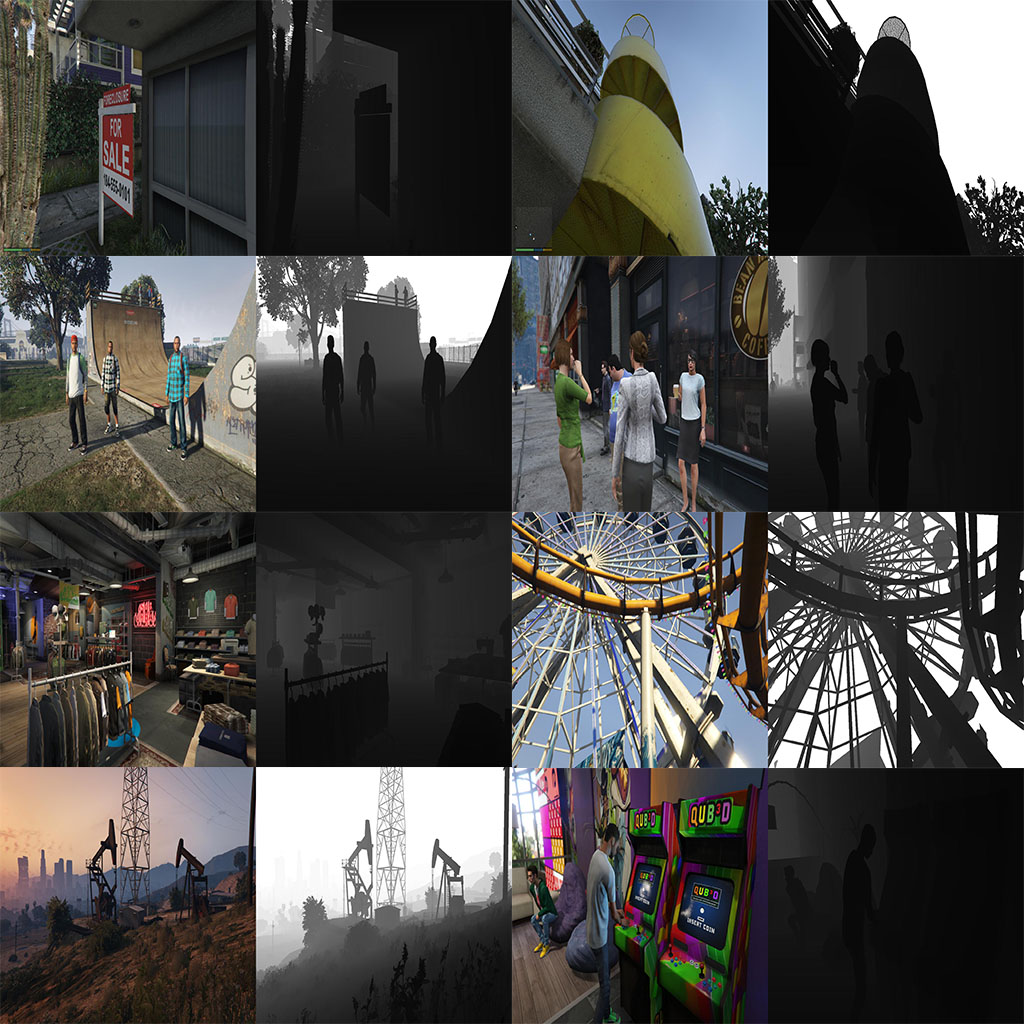}
    \caption{Dataset samples from different locations}
    \label{fig:dataset-examples-appendix}
\end{figure}

Fig \ref{fig:dataset-examples-appendix} shows a variety of different situations where we can extract depth maps in the wild to use as training data.

\begin{figure}[!ht]
\centering
\begin{subfigure}{.5\textwidth}
  \centering
  \includegraphics[width=0.95\linewidth]{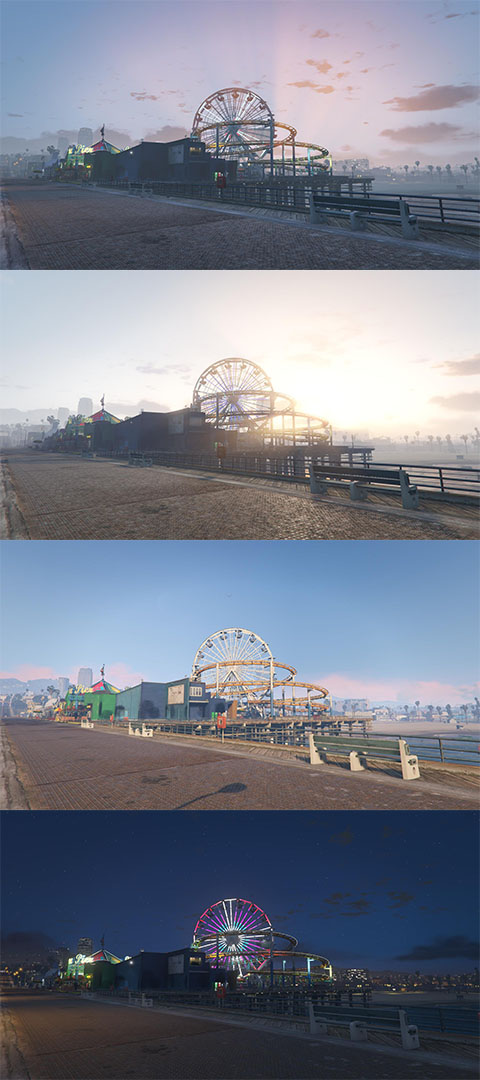}
  \caption{Day/Night Cycles}
  \label{fig:day-night-cycles}
\end{subfigure}%
\begin{subfigure}{.5\textwidth}
  \centering
  \includegraphics[width=0.95\linewidth]{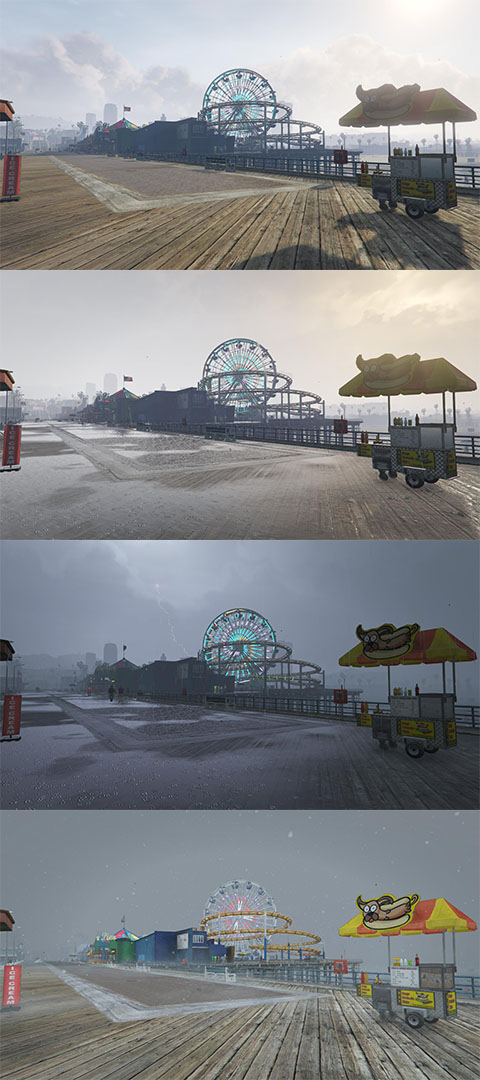}
  \caption{Weather Transitions}
  \label{fig:weather-transitions}
\end{subfigure}
\caption{Dynamic settings to generate more samples}
\label{fig:dynamic-settings-effects}
\end{figure}

Fig \ref{fig:dynamic-settings-effects} showcases different ways of acquiring training data to account for illumination and weather changes. By recording the transitions and using them as training data, we can help the model stay invariant to dynamic changes that don't affect depth estimation such as the sun and shadows, moving clouds, puddle reflections or day/night cycles.

\begin{figure*}[!ht]
    \centering
    \includegraphics[width=1.0\linewidth]{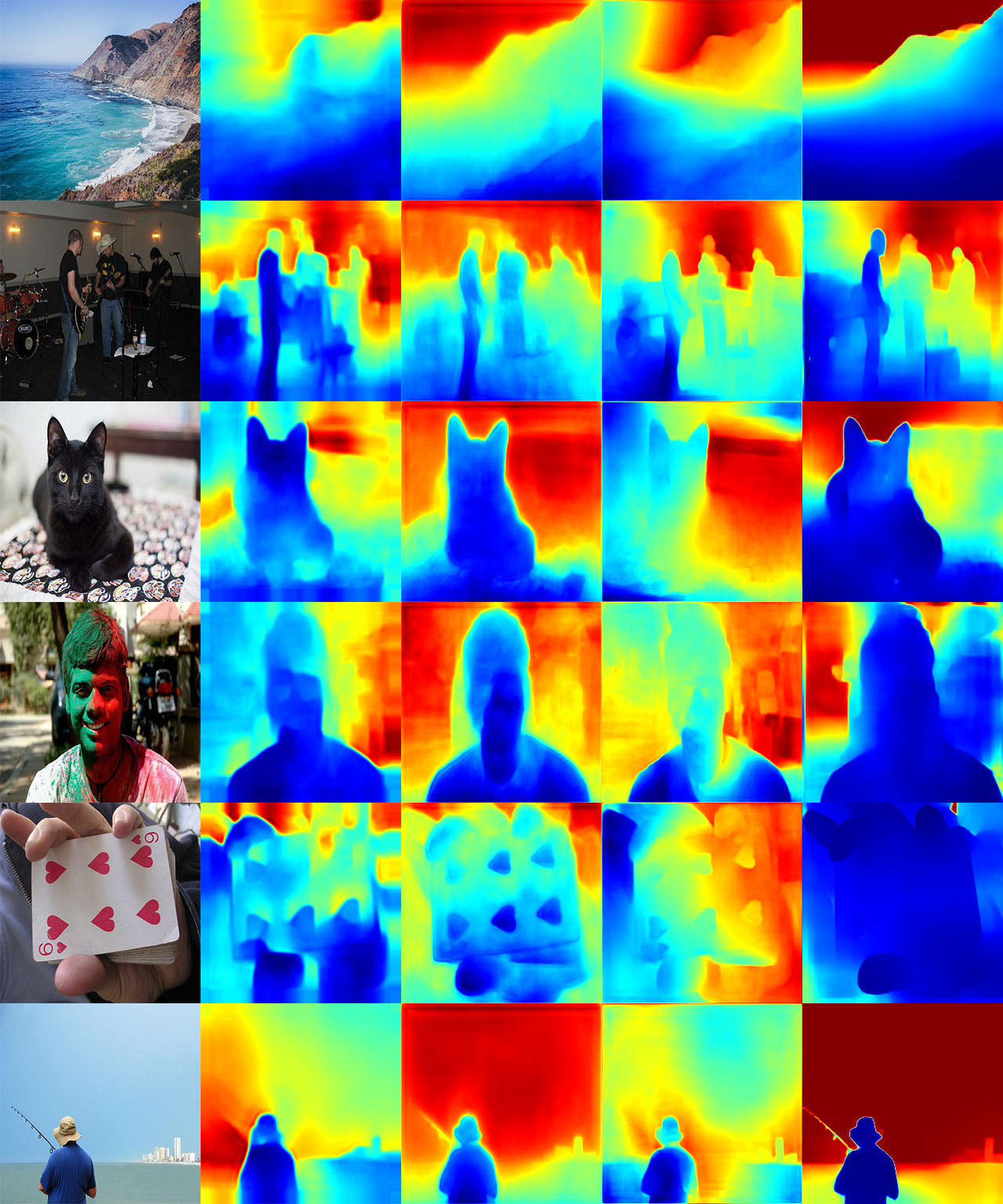}
    \caption{Qualitative Results on the DIW dataset. From left to right: Input Image, Eigen \cite{Eigen2016}, DIW (NYU\_DIW) \cite{Chen2016a}, DIW (Full) \cite{Chen2016a}, Our Approach}
    \label{fig:qualitative_results_extended}
\end{figure*}

Fig \ref{fig:qualitative_results_extended} compares results between \cite{Eigen2016}, \cite{Chen2016a} and our approach on the DIW dataset. As can be seen, our approach shows resistance to rapid changes in brightness intensity and illumination. Since we are not directly using skip-connections throughout our network, our model is less vulnerable to texture pattern phantoms which can be seen in the case of the pattern below the cat or the hearts on the card.

\clearpage
\bibliographystyle{unsrt}
\twocolumn
\bibliography{GamingForDepth}

\end{document}